# Machine Learning-based feasibility estimation of digital blocks in BCD technology

Francesco Daghero
Department of Control and Computer Engineering
*Politecnico di Torino*
10129, Turin, Italy
francesco.daghero@polito.it

Gabriele Faraone
Analog, Power & Discrete MEMS and Sensor R&D group
*STMicroelectronics*
20864, Agrate Brianza, Italy
gabriele.faraone@st.com

Michelangelo Grosso
Analog, Power & Discrete MEMS and Sensor R&D group
*STMicroelectronics*
10129, Turin, Italy
michelangelo.grosso@st.com

Daniele Jahier Pagliari
Department of Control and Computer Engineering
*Politecnico di Torino*
10129, Turin, Italy
daniele.jahier@polito.it

Nicola Di Carolo
Analog, Power & Discrete MEMS and Sensor R&D group
*STMicroelectronics*
20864, Agrate Brianza, Italy
nicola.dicarolo@st.com

Giovanna Antonella Franchino
Analog, Power & Discrete MEMS and Sensor R&D group
*STMicroelectronics*
20864, Agrate Brianza, Italy
giovanna.franchino@st.com

Dario Licastro
Analog, Power & Discrete MEMS and Sensor R&D group
*STMicroelectronics*
10129, Turin, Italy
dario.licastro@st.com

Eugenio Serianni
Analog, Power & Discrete MEMS and Sensor R&D group
*STMicroelectronics*
20864, Agrate Brianza, Italy
eugenio.serianni@st.com

*Abstract*—Analog-on-Top Mixed Signal (AMS) Integrated Circuit (IC) design is a time-consuming process predominantly carried out by hand. Within this flow, usually, some area is reserved by the top-level integrator for the placement of digital blocks. Specific features of the area, such as size and shape, have a relevant impact on the possibility of implementing the digital logic with the required functionality. We present a Machine Learning (ML)-based evaluation methodology for predicting the feasibility of digital implementation using a set of high-level features. This approach aims to avoid time-consuming Place-and-Route trials, enabling rapid feedback between Digital and Analog Back-End designers during top-level placement.

## I. Introduction

In the field of electronic Integrated Circuit (IC) design, the Physical Design (PD) of Analog and mixed-signal (AMS) ICs is often based on the "Analog-on-Top" approach. This approach involves instantiating all digital subsystems as intellectual property (IP) blocks and is particularly suitable when the analog components of the IC outnumber the digital ones.

In the Analog-on-Top approach, the top-level layout phases primarily focus on placing and interconnecting the analog components, while the digital portion is integrated into the remaining free silicon area of the chip. The placement of analog components on the die area is predominantly done with a manual and time-consuming methodology. This manual approach allows designers to explore a wide range of layout solutions based on desired performance, power consumption requirements, and constraints related to minimization of noise and variability effects. [1][2]

Once the area allocated for the digital logic has been specified, the standard way to verify both the placeability and routability of digital blocks, is to run the entire Place and Route flow with digital EDA tools, and then check the results of such run, to verify that timing, power and area requirements are met. In particular, the feasibility of the placement and routing of an input digital netlist depends on many design and technology-specific variables. Clearly, the designated placement area must be sufficiently large to fit the logic gates and interconnects derived from synthesis, while being strategically constrained to enhance cost-effectiveness. However, as the initial logic gates density (often expressed as row utilization) approaches the limits of the technology node, other factors become significantly relevant in addition to the number of metal layers used for power, clock, and signal routing. These factors present additional challenges to the placement and routing feasibility. For example, factors such as:

- the composition of the logic cells, either sequential or combinational,
- the regularity of the shape, which have an impact on routing congestion (e.g., a square or rectangular, regular shape is generally better than very thin boxes or L-shaped regions),
- the number and the position of pins (i.e., interfaces between the analog and the digital subsystem) along the shape edges,
- the properties of the clock tree in the circuit,

all contribute to the complexity of the placement and routing process. Experienced digital back-end physical design engineers can readily identify clear-cut unfeasible situations, while additional analysis and Place-and-Route (PnR) trials may be necessary to make a final determination and provide recommendations for improving placeability and routability, especially in less evident situations. This may involve actions such as relocating pins or modifying the allocated shape. In Analog-on-Top design of AMS circuits, where the digital logic is often confined to a small sub-area of the chip, this leads to **time-consuming iterations** between analog and digital design back-end engineering teams. Therefore, implementing an automated workflow with Machine Learning (ML) can significantly reduce the turn-around time (TAT) for such tasks (refer to Fig. 1).

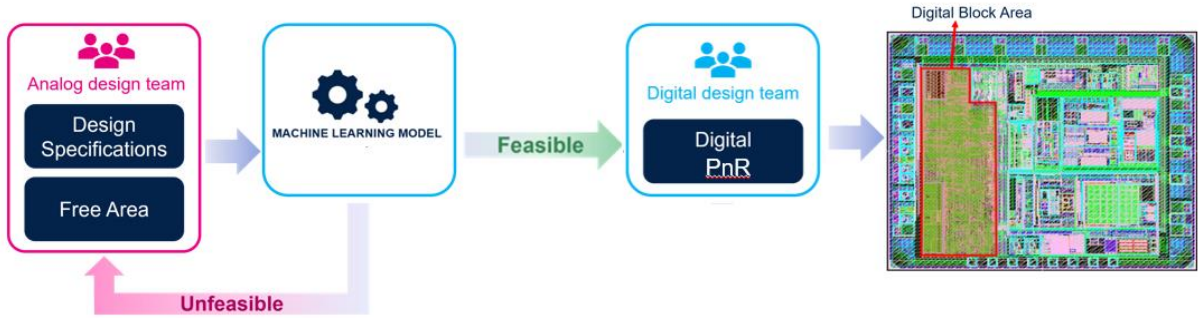

Fig. 1. Illustration of a Machine Learning aided Digital PnR feasibility estimation step within the digital design implementation flow of an AMS IC.

In our work, we propose an automated methodology based on a ML model that is able to assign a feasibility score to the whole Place and Route phase of a digital IP. This model takes into account high-level features of input netlist, expected layout characteristics, and technology information. The developed model offers the advantage of **easy implementation** in a top-level Analog placement context. The feasibility score provided by the model can assist the top-level Analog layout designer in determining the area to be allocated for the digital logic. This, in turn, **helps to reduce design time and costs**. Additionally, due to its simplicity, the model can be easily re-trained to get PnR feasibility estimation of digital IPs implemented in different technology nodes, providing the model with **good portability**.

This paper is organized as follows. Section II outlines the background and the context of the work; Section III details the methodology, while Section IV and V present the implementation and validation of our ML model for estimating the feasibility of the Place and Route (PnR) phase of digital designs implemented in ST-proprietary Bipolar-CMOS-DMOS (BCD8sp) technology. Section VI reports the application of the ML model to a real test-case. Finally, Section VII summarizes the conclusions of our work.

## II. Context

Numerous efforts have been proposed in the past to automate the physical design phase of AMS ICs [3][4][5], however these attempts can be considered effective only within their application domains and no physical design flows have been demonstrated yet to be universally valid for all classes of AMS circuits. Conversely, in digital design the physical implementation is highly automated by means of specialized flows and optimized standard cell libraries [6] being easily generalizable to all classes of digital circuits.

Machine learning (ML) and Artificial Intelligence (AI) techniques have become a valid asset in the Electronic Design Automation (EDA) industry since several years, helping to make cutting-edge design and verification tools: from synthesis to PnR, timing analysis, analog design, and circuit simulations [7][8][9]. Within such context, it is highly desirable to foster the development of a series of Electronic Design Automation (EDA) "scriptware" and methodologies, based on Artificial Intelligence (AI) and Machine Learning techniques that could enhance the productivity of Analog/Mixed-Signal (AMS) physical design allowing a full automation of the AMS Back-End (BE) flow.

This is the goal of the "**Analog Mixed-Signal Back-End Design Automation with Machine Learning and Artificial Intelligence Techniques" (AMBEATion)** Project, funded by the European Union within the Horizon 2020 research and innovation program under the Marie Skłodowska-Curie grant agreement No 101007730 [10][11], to which both companies (STMicroelectronics and Synopsys) and academia (Politecnico di Torino, Università di Catania, and Czech Technical University in Prague) collaborate. The project started in September 2021 and is expected to end in 2025. Within the AMBEATion framework, AI and ML algorithms can be introduced in at least three different steps of the Analog Mixed-Signal back-end implementation flow: at the top level analog schematic recognition, in the layout placement step and in the placement and routing (PnR) of Digital IPs. [12] This work specifically targets the latter step, by providing an automated methodology for assessing the feasibility of digital implementation within a constrained area, based on Machine Learning.

Previous approaches to feasibility estimation of digital PnR have primarily addressed the development of digital area estimators using probabilistic models [13][14] or high-level synthesis area estimation techniques [15][16]. However, there has been limited research on machine learning-based methods for estimating the area of digital hardware design in past literature. Specifically, the developed ML models have focused on either directly estimating the area of digital hardware components from specifications [17][18] or predicting circuit performance metrics after the global route phase of digital Place and Route (PnR) [19].

## III. METHODOLOGY

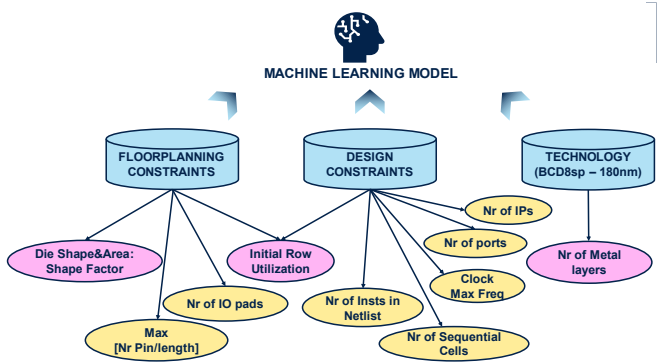

Fig. 2. Illustration of input features of a typical PnR flow. Highlighted in pink the features actually selected for the training of the DT model.

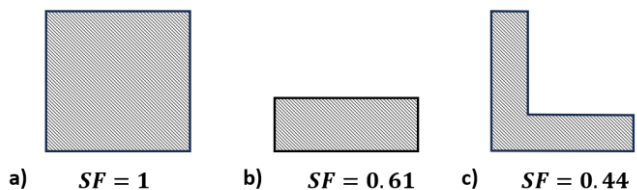

Fig. 3 Shape Factor for different (die) shapes: unit square with Area 1 a), narrow rectangle b) and L-shape c). Note that b) and c) have the same Area (A=0.4375) but a different SF value. In c) SF is lower due to the presence of > 4 edges and an elbow which may eventually affect the routability during PnR phase especially in case of digital designs with high row utlization.

From a general perspective, a ML model developed for predicting the area occupied by digital logic after the PnR phase, or for determining the feasibility of the entire PnR phase of a digital circuit within an assigned area, must identify the relevant input features that significantly influence key metrics (such as area, power, and timing performances) of the physical implementation of a digital design. The physical implementation of a digital IP is, however, a highly complex problem that depends on numerous variables and, therefore, requires a careful selection of the model's input features. In Figure 2, we have identified the features that typically have the most significant impact on digital design metrics after Place and Route. These features include:

- the layout (die) area and shape;
- the number of instances (combinatorial or sequential cells) in the input netlist to be placed in the digital core;
- the number of macro blocks (e.g., analog IPs or memories – if any) to be included in the digital core;
- the initial row utilization, i.e., the ratio of the sum of the area of all instances in the input netlist over total *placeable* area in the layout;
- the number of sequential cells in the design (flip-flops, latches and integrated clock gating);
- the maximum frequency of clock for clocked designs;
- the total number of in/out ports and the number of I/O cells (pads, if included in the digital netlist);
- the (numerical) density of pins expressed as the max value of the ratio between number of pins along an edge and the edge length;
- technology information, such as the number of available routing layers (including those used for power supply connections).

Another aspect to bring into consideration when choosing the most appropriate ML model is the dataset size used for training. The dataset considered here consists of different digital design databases containing information (value of input parameters, and performance metrics) of their Place and Route Phase. For small datasets simple algorithms, such as logistics regression or decision trees [20] (DT) may perform better, whereas larger dataset may require more complex algorithms like Random Forests [21], Support Vector Machine [22] or Multilayer perceptron [23].

## IV. MODEL IMPLEMENTATION

In our work we opted to implement the ML model to estimate the feasibility of Digital PnR with a decision tree. The DT was trained in a supervised manner using data extracted from EDA tool logs of both successful and failed PnR runs from previous designs. While EDA tool logs were commonly available for circuits that reached the tape-out, obtaining data for intermediate unsuccessful PnR trials is less common. The DT approach, therefore, was selected for its effectiveness and simplicity, particularly when working with limited training data allowing also an easy qualitative interpretation of the model results.

All the training data were selected from digital designs implemented in Bipolar-CMOS-DMOS (BCD8sp) technology, which offers a wide variety of projects at STMicroelectronics, providing a reasonably good variability in the training data. A total of 47 different digital designs were collected, and Table I illustrates the designs divided according to the digital netlist area, giving an idea of their complexity. For some designs, both failing and non-failing PnR implementation logs were available, while in other cases, the entire digital PnR flow was rerun to obtain failing and successful PnR logs for the same design. This resulted in a dataset of 96 samples, which was used to train the decision tree model.

TABLE I.

| Area (A) [mm²] | Designs |
|---|---|
| 0 ≤ A < 0.5 | 45.54 % |
| 0.5 ≤ A < 1 | 26.73 % |
| 1 ≤ A < 4 | 20.79 % |
| A ≥ 4 | 6.94 % |

In the choice of the input features of the DT model we observed that die area and shape strongly affect both the placement and routability of the digital PnR. The feasibility of the placement is governed by the value of the initial row utilization, whereas the routability depends on the routing congestion, which is mainly influenced by the number of instances in the netlist, the number of routing layers, and the die shape. Therefore, in determining the most relevant input feature for the model, we chose a subset among the list above including only:

- the number of routing layers,
- the initial row utilization and
- the layout area and its shape.

Regarding the layout area and its shape, we observe that their effect on the routability and placeability of a digital design is strongly intertwined. For instance, when the number of instances in the netlist and the number of available routing layers are held constant, the signal routing of two identical digital designs allocated in two regions with equal area (and therefore having the same initial row utilization) may be more or less feasible based on the shape of the layout region. A square-like shape generally presents fewer routing issues compared to irregular shapes characterized by many edges and/or elbows. To describe the combined effect of shape and area in the feasibility of digital PnR, a **Shape-Factor** was introduced. This number gives a score to the shape of the layout region, based on its similarity to a perfect square and it ranges from 0 to 1. The more similar the die-shape is to a square, the closer it is to 1, as illustrated in Figure 3. It is computed by dividing the normalized perimeter $\tilde{P}$ by the normalized area $\tilde{A}$ of the die (both quantities obtained considering the polygon sides normalized with respect to the longest one). This is done to obtain measurements as independent as possible from the absolute geometrical data. Accounting that the area of a unitary square is 1 and its perimeter 4, the Shape-Factor SF is defined as:

$$SF = \frac{4\tilde{A}}{\tilde{P}} \qquad (1)$$

In Figure 2, the three input features selected for training the decision tree are highlighted in pink. During the training phase of our decision tree, we chose to limit the maximum depth of the tree to 3 and utilized balanced class weights. The constraint on maximum depth was essential to prevent extreme overfitting, while the use of balanced class weights was crucial to address the limited number of failing design samples in our training dataset.

The outcome of the ML algorithm training is shown in Figure 4. It represents a decision tree where, going from a certain node to the left child means that the condition reported in the first line of each node is held. Otherwise, the right child is selected. The decision tree tries to make a prediction on the class (successful PnR or unsuccessful PnR) to which an input digital design allocated in certain die-area belongs by testing the value of the three predictor variables discussed in the previous paragraph at different branches.

In the training phase of the tree, the algorithm tries to determine which variable (among the three chosen) will split the data so that the underlying child nodes are most homogeneous or pure (i.e., completely belonging to a class or another). The variable selection criterion of the algorithm is performed via the Gini Index, which basically measures the probability of a random data being misclassified. Thus, the algorithm tries to minimize (make close to 0) the Gini index at each node. As can be seen in figure 4 the leaf nodes of the tree have the lowest Gini index value, meaning that the likelihood of misclassification of the data is lower. Reported in each node is also the percentage of training samples reaching the node and the probability of belonging to one of the two classes in square brackets: [not feasible, feasible]. Note that the colors of the nodes are also chosen according to these probabilities: successful (blue) or unsuccessful (orange) digital PnR implementation. Therefore, moving down the tree we can infer that a digital design allocated in a die-area having a shape characterized by a shape-factor lower than 0.7 (e.g., a narrow rectangle or a polygon with number of edges > 4) implemented with a BCD8sp technology option having more than 3 routing layers, and having an initial row utilization lower than 88.5% is feasible with a score of 56.6% (see nodes highlighted in red in figure 4). This estimation, made from the DT model, is based on 29.5 % of the training data. Conversely a digital design that has to be allocated into a more square-like area (shape-factor higher than 0.7), implemented in BCD8sp technology with more than 3 routing layers and having an initial row utilization lower than 83.6% is always feasible with a score of 100% (nodes highlighted in green in figure 4). Such inference is based accounting that such choice of input features was based on 41.1 % of the training data.

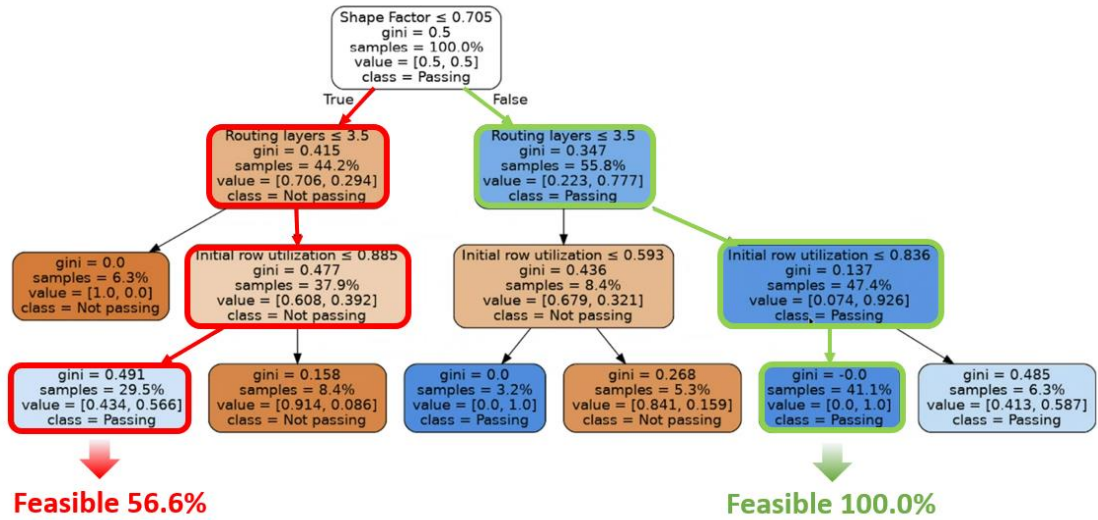

Fig. 4 Graphical representation of the decision tree ML model used for the feasibility evaluation of Digital PnR. Highlighted in green and red are two possible Digital PnR feasibility inferences as discussed in the text.

## V. VALIDATION

To evaluate the accuracy of the DT tree model, we utilized the *Leave-One-Out cross-validation approach*. This involved training the model on all samples except one, which served as the test set. The process was repeated until all projects with at least one failing design (fold) were utilized as the test set, which occurred 8 times in our case. Table II reports the test balanced accuracy of each fold. With the aforementioned approach, we achieve 76% average balanced accuracy, however with a variability of $\pm 13\%$, as some folds include only few samples.

Results in table II with roughly 50% balanced accuracy may be ascribed to folds in which the test set is affected, at least, by a combination of the following factors:

- Extreme differences between the designs belonging to test and training sets;
- Small test sets size compared to the one of the training data;
- Predictions on unfeasible designs in the test set (from a model trained on unbalanced data) often classified as false negatives, reducing the overall accuracy.

Although the variability on model's average accuracy is 13%, qualitative analysis (Figure 4) indicates that it learns reasonable rules in most cases. This suggests a potential improvement with a larger dataset or, eventually, incorporating additional input features into the model. In particular, the ML model shows proficient results when is applied on the class of digital IPs which are designed in our product-division (see section V). However, it is expected that some problems may occur when we evaluate the PnR feasibility of a digital IP which has less-common layout features. For example, when we consider a design to be routed in less than 3 metal layers and with a shape factor less than 0.7. The evaluation of such design might produce naïve conclusion because of the lack of training points (6.3% of the points).

Finally, we report that the **execution time** of such decision tree models is minimal, requiring less than 2 ms for training and **1.5 ms** for inference.

TABLE II.

| Fold | Balanced Accuracy |
|---|---|
| 1 | 75% |
| 2 | 100% |
| 3 | 75% |
| 4 | 75% |
| 5 | 80% |
| 6 | 50% |
| 7 | 75% |
| 8 | 75% |
| **Mean $\pm$StdDev** | $(76 \pm 13)$ % |

## VI. APPLICATION

As a real case of application, our validated decision tree model was implemented in the back-end design phase of a power-management IC (PMIC) to be used in a real product addressed to the memory storage market. The product was designed in BCD8sp technology node in Analog-on-Top approach and contains 10 different digital modules. The digital feasibility estimation model was used to provide PnR feasibility estimations on the three biggest digital IPs and the following table shows the initial predicted feasibility scores.

TABLE III.

| Initial Row Utilization (%) | Nr. Routing Layers | SF | PnR Feasibility score (%) |
|---|---|---|---|
| 87.69 | 4 | 0.7823 | 58.66 |
| 94.07 | 4 | 0.4400 | 8.64 |
| 69.12 | 4 | 0.8936 | 100 |

It is worthwhile noting that the runtime of the PnR flow (from the import design step to the post-routing optimization step) for the three digital IPs considered in the table above was about 30min for the first two digital modules and about 3 hours for the latter. Normally, in all the three situations different PnR flow iterations are necessary to evaluate the best die-area compromise for a feasible placement and routing of the digital modules without overlap errors and design rule (DRC) check violations. The application of our decision tree model, alternatively, provided an immediate (3ms) response on the feasibility of the PnR saving considerable time in the feedback between Analog and Digital back-end designers. In particular, the second digital IP considered in the table had a PnR feasibility score of 8.64%. This feedback was used to address better the hardware implementation of these digital IPs by removing spare cells and Design for Testability added logics (test-point insertions) so as to reduce the initial row utilization value to converge toward a feasible PnR flow in the allocated area.

## VII. CONCLUSION

Here we have shown that a suitably trained decision tree ML model is able to provide automatically and in **short time a prediction** on the feasibility of PnR of digital blocks within an allocated area. This aspect is particularly relevant from the point of view of an **AMS back-end design**, where automating decisions in a top-level chip placement context, can help to significantly **reduce time-to-market and turnaround time**. In this respect, getting a feasibility estimation of the digital PnR implementation earlier in the design flow and with the use of automated procedures can help also to carry out meaningful evaluations of the minimum silicon area that must be allocated for the digital portion of an AMS IC design targeting the reduction of **silicon manufacturing costs.**

The methodology illustrated here paves the way toward an automated, machine learning based AMS BE IC design flow, in which a machine-learning based digital PnR area-feasibility evaluation is an integral of an industrial design methodology.